%% file: main.tex
\documentclass[10pt,twocolumn,letterpaper]{article}

\usepackage[pagenumbers]{cvpr} 

\usepackage[utf8]{inputenc} 
\usepackage[T1]{fontenc}    
\usepackage{amsmath}
\usepackage{url}            
\usepackage{booktabs}       
\usepackage{amsfonts}       
\usepackage{nicefrac}       
\usepackage{microtype}      
\usepackage{xcolor}         
\usepackage{varwidth}
\usepackage{graphicx}
\usepackage{float}

\usepackage{tikz}
\usepackage{tikz,tikz-3dplot}
\usetikzlibrary{calc,fit,spy}

\newcommand{\bbf}{\mathbf{f}}
\newcommand{\bbg}{\mathbf{g}}
\newcommand{\bbx}{\mathbf{x}}

\newcommand{\ours}{NODE-Flow}

\usepackage{multirow}
\usepackage{adjustbox}

\definecolor{cvprblue}{rgb}{0.21,0.49,0.74}
\usepackage[pagebackref,breaklinks,colorlinks,allcolors=cvprblue]{hyperref}

\def\paperID{7} 
\def\confName{CVPR}
\def\confYear{2025}

\title{Learning Optical Flow Field via Neural Ordinary Differential Equation}

\author{
Leyla Mirvakhabova~~~
Hong Cai~~~
Jisoo Jeong~~~
Hanno Ackermann~~~
Farhad Zanjani~~~
Fatih Porikli~~~ 
\smallskip
\\
Qualcomm AI Research$^{\dagger}$~~~
\\
\smallskip
{\tt\small\{lmirvakh, hongcai, jisojeon, hackerma, fzanjani, fporikli\}@qti.qualcomm.com \vspace{-12pt}}
}

\begin{document}
\maketitle
\begin{abstract}

Recent works on optical flow estimation use neural networks to predict the flow field that maps positions of one image to positions of the other. These networks consist of a feature extractor, a correlation volume, and finally several refinement steps. These refinement steps mimic the iterative refinements performed by classical optimization algorithms and are usually implemented by neural layers (e.g., GRU) which are recurrently executed for a fixed and pre-determined number of steps. However, relying on a fixed number of steps may result in suboptimal performance because it is not tailored to the input data.

In this paper, we introduce a novel approach for predicting the derivative of the flow using a continuous model, namely neural ordinary differential equations (ODE). One key advantage of this approach is its capacity to model an equilibrium process, dynamically adjusting the number of compute steps based on the data at hand. By following a particular neural architecture, ODE solver, and associated hyperparameters, our proposed model can replicate the exact same updates as recurrent cells used in existing works, offering greater generality. 

Through extensive experimental analysis on optical flow benchmarks, we demonstrate that our approach achieves an impressive improvement over baseline and existing models, all while requiring only a single refinement step. 

\end{abstract}

\begin{figure*}[h!]
\begin{center} 
\includegraphics[width=\textwidth]{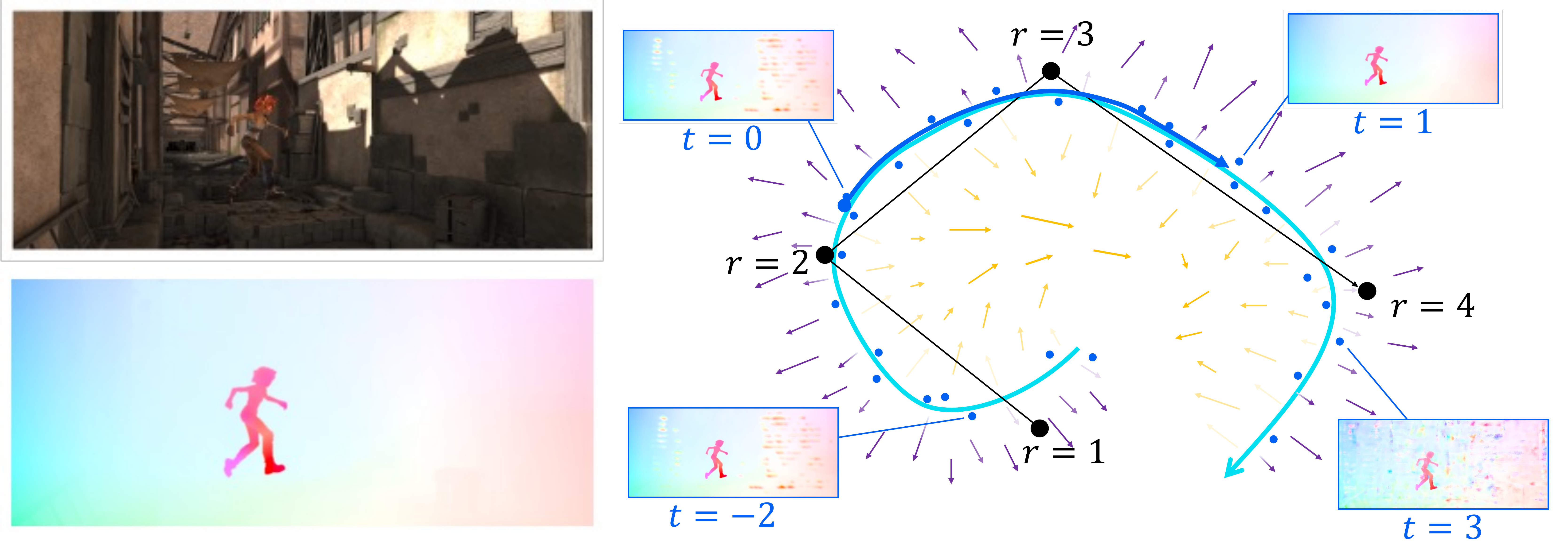} %
\caption{
Upper left: source image. Lower left: corresponding GT flow. Right: ODEs can represent a vector field, thus the time scale parameter of the neural ODE solver generates points along the path (both in blue) along the GT curve (cyan). Compared to the fixed number of steps of an RNN (black lines and points), the neural ODE does not overshoot because its termination criterion is data-driven.}
\label{exp:schematic}
\end{center}
\end{figure*}

\section{Introduction}
\label{sec:intro}
{\let\thefootnote\relax\footnotetext{{
\hspace{-6.5mm} $\dagger$ Qualcomm AI Research is an initiative of Qualcomm Technologies, Inc.}}}
Optical flow (\textbf{OF}) is a fundamental challenge in computer vision aiming to estimate a dense displacement field between consecutive video frames. By assuming appearance consistency, where pixels at corresponding locations in these frames share similar appearances, traditional approaches tackle this by solving a differential equation or optimization problem to derive the displacement field.

In recent years, there has been a notable progress in this field, particularly with the emergence of deep learning based models, such as recurrent neural networks (RNNs). Among these, RAFT~\cite{teed2020raft} stands out for its significant accuracy improvement and strong generalizability as compared to previous optimization- and learning-based models. It utilizes a gated recurrent unit (GRU) cell for iterative flow estimation updates, resembling an optimization process. This is a critical component of RAFT and this approach has been widely embraced by the latest state-of-the-art optical flow models, including  FlowFormer~\cite{huang2022flowformer}.

While this recurrent approach with GRU cells yields improved performance, it comes with several limitations. Firstly, such a design mandates a discrete setting for the refinement process, which constrains learning precision, especially in scenarios with complex underlying dynamics. Moreover, from a practical standpoint, employing a recurrent neural module requires a prescribed number of iterations. In existing optical flow literature, these iteration counts are defined manually or through grid search. However, such counts may vary between training and testing datasets or due to discrepancies in image-to-image motion. Consequently, there is a lack of a principled way to determine the optimal number of iterations, particularly when deploying the model on new, unseen data.

Another known drawback of RNNs relates to backpropagation-through-time (BPTT) that imposes the difficulties in converging to local optima ~\cite{werbos1990backpropagation, bengio1994learning}. In this setting, the optimization is done through unfolded RNN iterations, complicating the modeling of long-term dependencies. 

In this study, we take a step back and revisit deep learning based OF estimation, questioning the efficacy of current algorithms in adequately capturing OF dynamics. Starting with the Taylor-expansion of the color or brightness constancy assumption, conventional OF estimation algorithms substitute the temporal derivatives by the pixel-to-pixel differences. Using recurrent layers, on the other hand, necessitates fixing the number of iterations, thereby introducing aforementioned limitations. In contrast, we advocate for approximating these dynamics with neural ordinary differential equations, which offer superior capabilities in modeling time-varying processes. 


Unlike previous approaches that use either a GRU cell or a transformer architecture to iteratively compute the flow updates, we propose to predict the derivative of a latent flow, which is subsequently decoded to update the flow estimate. This is achieved by integrating a neural ordinary differential equation layer into our proposed architecture. This neural ODE layer represents a specific instance of an implicit layer tailored to effectively model temporal dynamics. As opposed to GRUs, our methods involves solving a differential equation on each step until convergence, with the flexibility to adjust the number of integration steps within the solver as needed.

In the process of solving neural ordinary differential equations (ODEs), the solver operates iteratively, but this process is fundamentally different from the standard iterative optical flow estimation. Typically, an iterative optical flow model refines its input, adding extra computational load. In contrast, the proposed method avoids recalculating the input and works solely with the initial approximation. Similarly, the backpropagation process differs: in baseline optical flow, the gradient is computed at each iteration that may potentially lead to BPTT, whereas a black box differential solver in Neural ODEs requires only constant memory and does not suffer from BPTT-related issues.


This allows for finer-grained learning of the optical flow field, resulting in significantly improved accuracy of the estimated optical flow, as demonstrated by our experimental results. With our proposed approach, there is no need to prescribe a fixed iteration number. Instead, we define a stopping criterion that allows the number of optimization steps to be adaptive based on the current data. Furthermore, any potential excessive memory consumption  due to a large number of steps can be mitigated by backward-mode integration during backpropagation. 

Additionally, we offer an interpretation of Neural ODEs as a broader extension of GRU-based updates. As detailed in \cite{de2019gru}, a GRU-ODE is introduced as a continuous-time variant of the GRU architecture, suggesting that Neural ODEs can represent the same dynamics learned by GRUs as a special case.\footnote{One can easily define such a model by choosing a solver with a fixed step update and maintaining the same number of refinement iterations as in the baseline GRU model.} Thus, opting for a Neural ODE as refinement block presents a more versatile framework for optical flow estimation, capable of capturing more complex dynamics.


In this paper, we select  GMFlow~\cite{xu2022gmflow} as a baseline model for its reported high accuracy across various optical flow estimation benchmarks. We demonstrate significant performance enhancements by substituting its GRU-based update module by implicit updates derived from the Neural ODE layer.

The main contributions of this paper are summarized as follows:
\begin{itemize}
    \item We introduce modeling of image-to-image temporal dynamics by ordinary differential equations.
    \item 
    In the proposed framework, the number of internal steps is guided by the prediction accuracy. On each step of the solver, we are approximating a solution of the differential equation, thus the total number of refinement steps in the optical flow model itself can be fixed to one. 
    \item The experimental results show better accuracy than baseline models that typically require multiple refinement stages.
\end{itemize}


\section{Related Work}
\label{sec:related}
\subsubsection*{Optical Flow Estimation} 
Optical flow is based on a brightness or color constancy assumption whose first-order Taylor-expansion yields the famous optical flow. Early, classical approaches solve for the OF either globally or locally, which are later combined~\cite{farneback2003two,bruhn2005lucas,horn1981determining,lucas1981iterative}. More recently, neural approaches like FlowNet~\cite{dosovitskiy2015flownet} propose task-specific correlation volumes, i.e.,  correlation maps between the two images on multiple resolutions, which have been used by many following works~\cite{ranjan2017optical, sun2018pwc, sun2019models, hui2018liteflownet, hui2019lightweight, yang2019volumetric}. 

To mimic the iterative optimization performed by classical algorithms, recurrent layers are proposed in RAFT~\cite{teed2020raft}. This algorithm uses 
GRU cells to refine the estimate in the last stage ~\cite{shi2015convolutional,liang2015recurrent,chung2014empirical}. This idea has since been used in many other algorithms~\cite{zhang2021separable, xu2022gmflow, jeong2022imposing, huang2022flowformer}. In GMFlow~\cite{xu2022gmflow}, the correlation volume is replaced by a differentiable matching on a single resolution. Although GMFlow originally uses a single step in the refinement module, the follow-up work of GMFlow+~\cite{xu2023unifying} demonstrates improved performance when using multiple refinement steps. Works like GMFlow, PerceiverIO~\cite{jaegle2021perceiver} and FlowFormer~\cite{huang2022flowformer} are based on transformer architectures~\cite{vaswani2017attention}. FlowFormer also uses multiple refinement steps. Recently, a solution based on diffusion has been proposed in~\cite{saxena2024surprising}.

\subsubsection*{Implicit Models} Models using implicit layers have been proposed for OF estimation~\cite{bai2022deep}. Implicit layers locate some fixed point via root finding~\cite{huang2021textrm}. They have been applied to 3D surface fitting~\cite{michalkiewicz2019implicit}, flow estimation~\cite{smith2022physics}, and solving boundary value problems~\cite{sitzmann2020implicit}. 
Due to the variable number of iterations, implicit methods can be more accurate than methods that rely on a pre-determined number of refinements.

\subsubsection*{Neural Ordinary Differential Equations (NODE)} This method also relies on an implicit layer~\cite{chen2018neural}. 
Differently from other implicit methods, an ordinary differential equation (\textbf{ODE}) is modeled, thus they parameterize the derivative of some quantity. 
Solvers, for instance based on Runge-Kutta methods, are used to determine the solution of the ODE by evaluating its integral. 
In the paper~\cite{chen2018neural}, the right-hand side of the ODE in Eq.~\eqref{eq:ode} is parameterized by a neural network. 
NODEs have all the advantages of other implicit methods while being more accurate and robust~\cite{hanshu2019robustness} when used to model time-varying differential processes~\cite{rubanova2019latent}. To reduce errors during reverse mode integration, gradient checkpointing has been proposed by~\cite{zhuang2020adaptive}.


\section{Method}\label{sec:method}


\input{method}
\section{Experiments}\label{sec:exp}


 \begin{figure*}[t]
 \begin{center}$
 \centering
 \begin{tabular}{c c c c}
 \hspace{-3mm} \rotatebox{90}{\quad \text{\small{\textbf{Overlay}}}} & 
 \hspace{-5mm} \includegraphics[width=4.5cm]{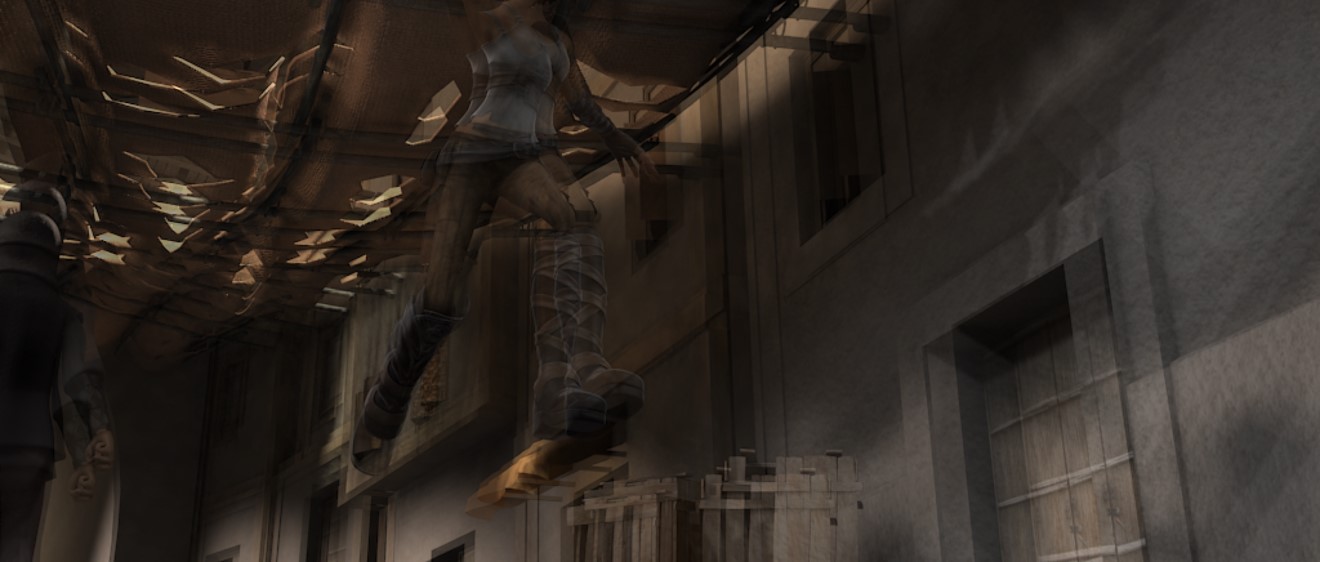} &
 \hspace{-5mm} \includegraphics[width=4.5cm]{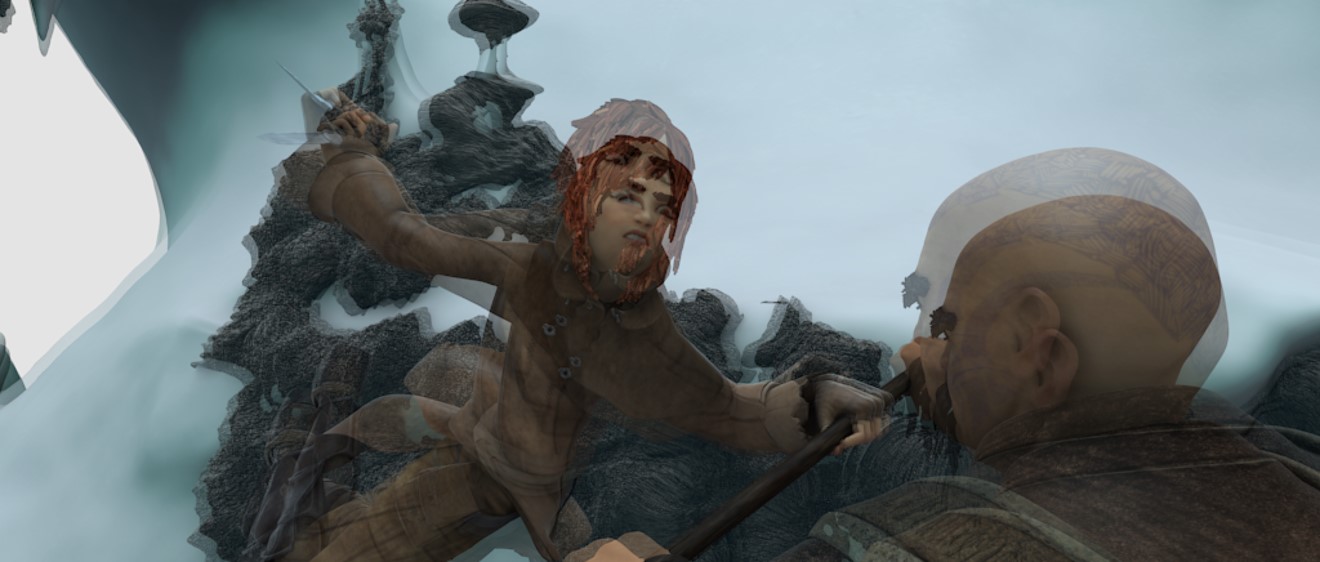} & 
\hspace{-5mm} \includegraphics[width=4.5cm]{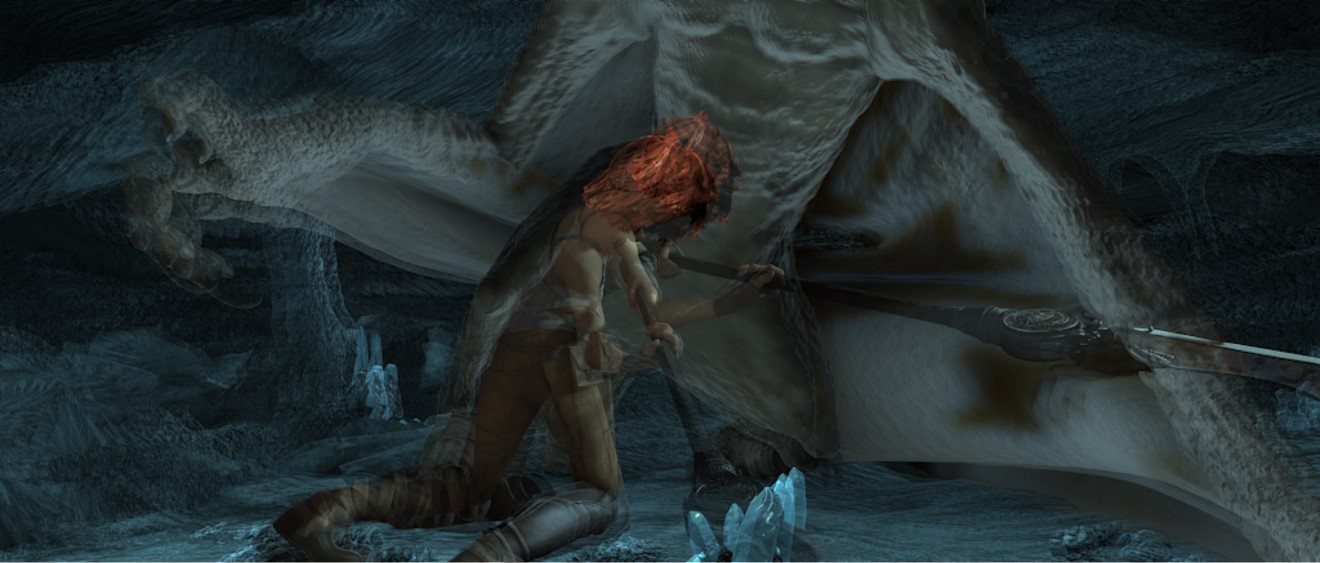} 
 \\
 \vspace{-2mm}
\hspace{-3mm} \rotatebox{90}{\qquad \text{\small{\textbf{GT}}}} & 
 \hspace{-5.5mm} 
 \begin{tikzpicture}[spy using outlines={rectangle,red,magnification=3, size=1.6cm, connect spies}] \node {\includegraphics[trim=0 0 0 0, clip, width=3.7cm, trim={0 0 0 0},clip]{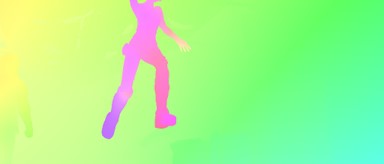}};
\spy on (-0.35,0.5) in node [left] at (2.65,0.0);\end{tikzpicture}
 & \hspace{-6mm} 
  \begin{tikzpicture}[spy using outlines={rectangle,red,magnification=3, size=1.6cm, connect spies}] \node {\includegraphics[trim=0 0 0 0, clip, width=3.7cm, trim={0 0 0 0},clip]{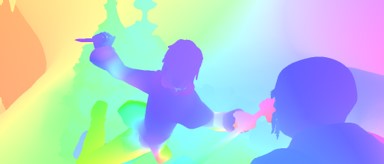}};
\spy on (-0.9,0.4) in node [left] at (2.65,0.0);\end{tikzpicture}
 & \hspace{-6mm} 
  \begin{tikzpicture}[spy using outlines={rectangle,red,magnification=3, size=1.6cm, connect spies}] \node {\includegraphics[trim=0 0 0 0, clip, width=3.7cm, trim={0 0 0 0},clip]{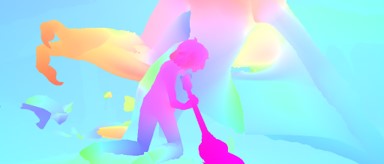}};
\spy on (0.3,-0.5) in node [left] at (2.65,0.0);\end{tikzpicture}
 \\
  \vspace{-2mm}
 \hspace{-3mm} \rotatebox{90}{\quad  \text{\small{\textbf{GMFlow}}}} & 
 \hspace{-6mm} 
 \begin{tikzpicture}[spy using outlines={rectangle,red,magnification=3, size=1.6cm, connect spies}] \node {\includegraphics[trim=0 0 0 0, clip, width=3.7cm, trim={0 0 0 0},clip]{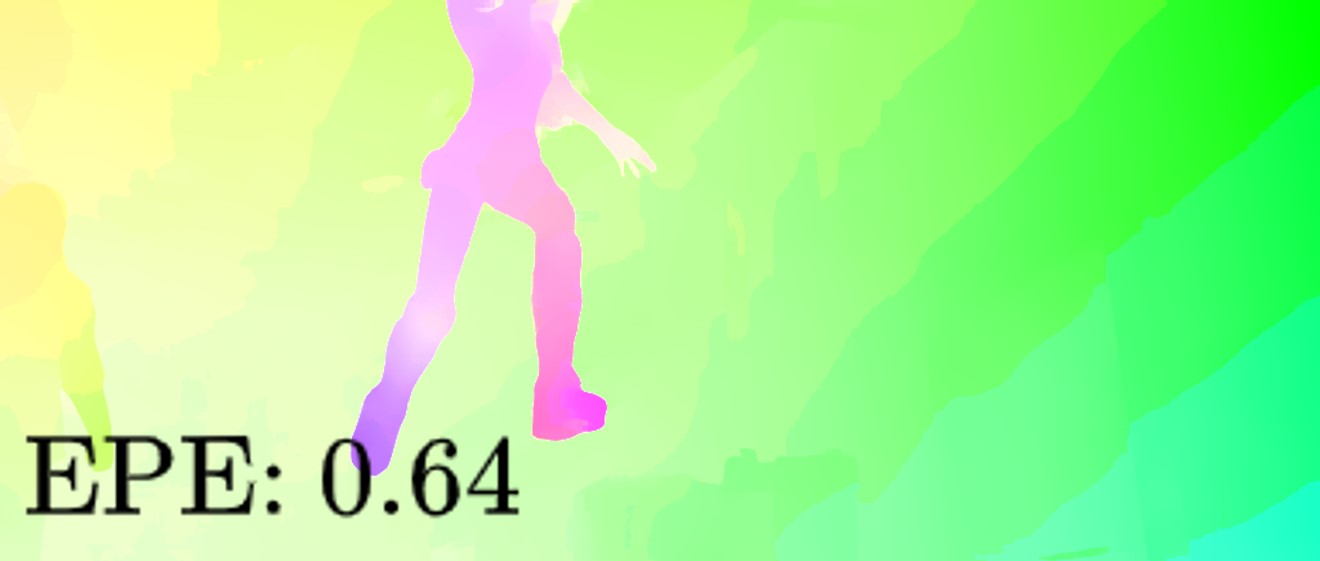}};
\spy on (-0.35,0.5) in node [left] at (2.65,0.0);\end{tikzpicture}
 & \hspace{-6mm} 
  \begin{tikzpicture}[spy using outlines={rectangle,red,magnification=3, size=1.6cm, connect spies}] \node {\includegraphics[trim=0 0 0 0, clip, width=3.7cm, trim={0 0 0 0},clip]{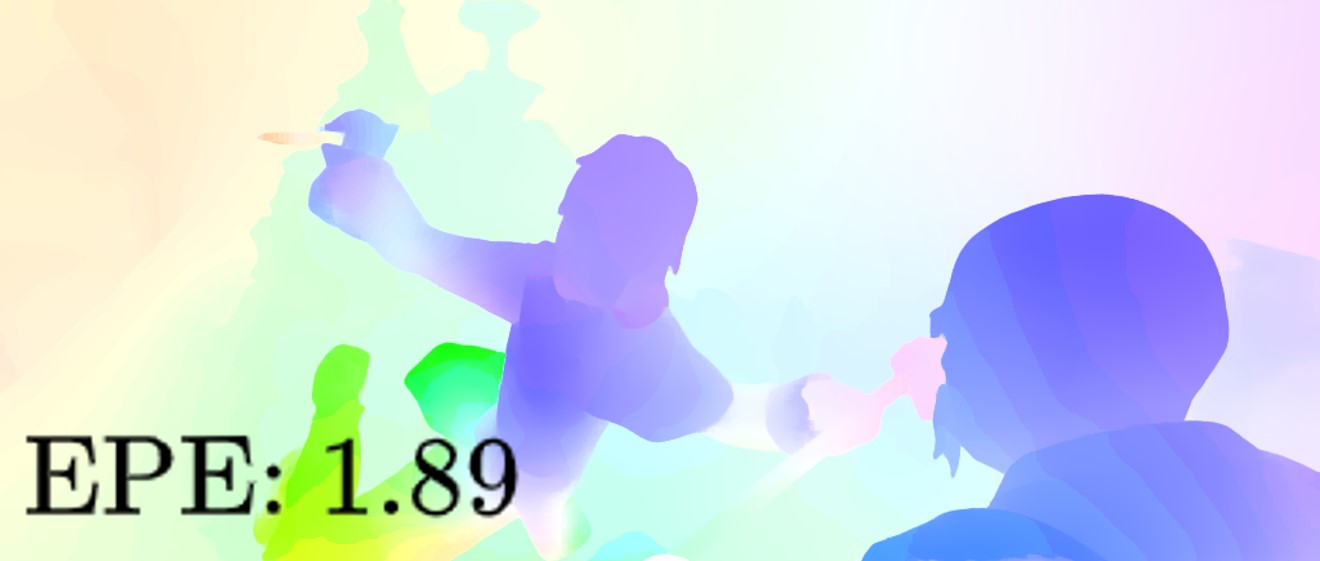}};
\spy on (-0.9,0.4) in node [left] at (2.65,0.0);\end{tikzpicture}
 & \hspace{-6mm} 
  \begin{tikzpicture}[spy using outlines={rectangle,red,magnification=3, size=1.6cm, connect spies}] \node {\includegraphics[trim=0 0 0 0, clip, width=3.7cm, trim={0 0 0 0},clip]{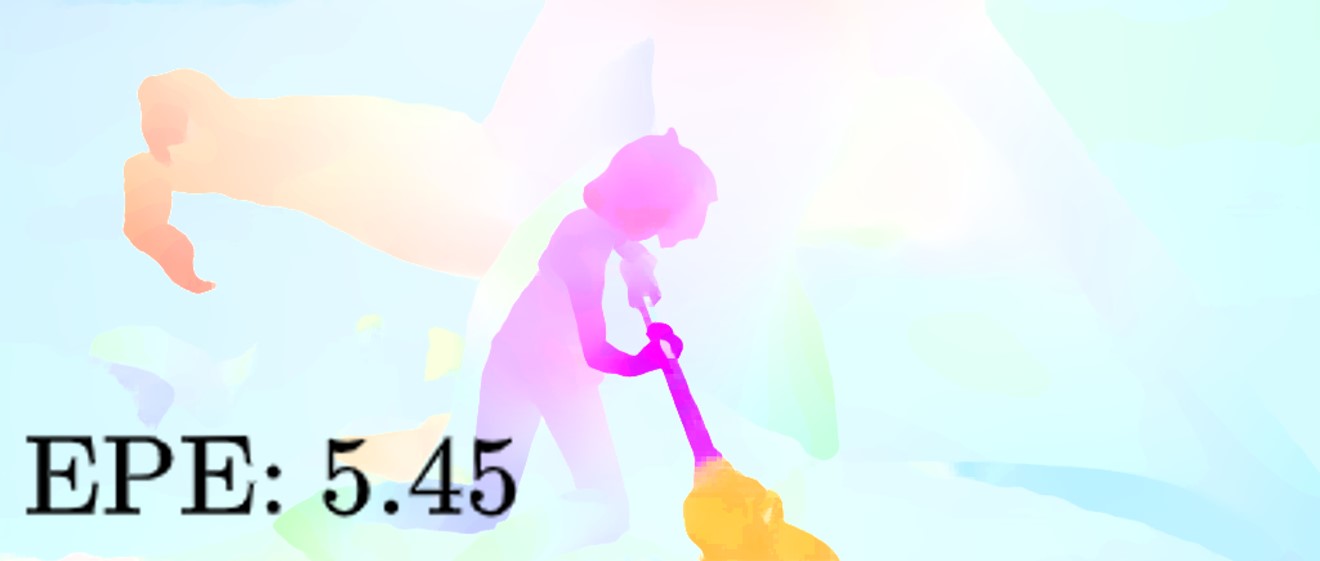}};
\spy on (0.3,-0.5) in node [left] at (2.65,0.0);\end{tikzpicture}
 \\
  \vspace{-2mm}
 \hspace{-3mm} \rotatebox{90}{\ \ \quad  \text{\small{\textbf{Ours}}}} & 
 \hspace{-6mm} 
 \begin{tikzpicture}[spy using outlines={rectangle,red,magnification=3, size=1.6cm, connect spies}] \node {\includegraphics[trim=0 0 0 0, clip, width=3.7cm, trim={0 0 0 0},clip]{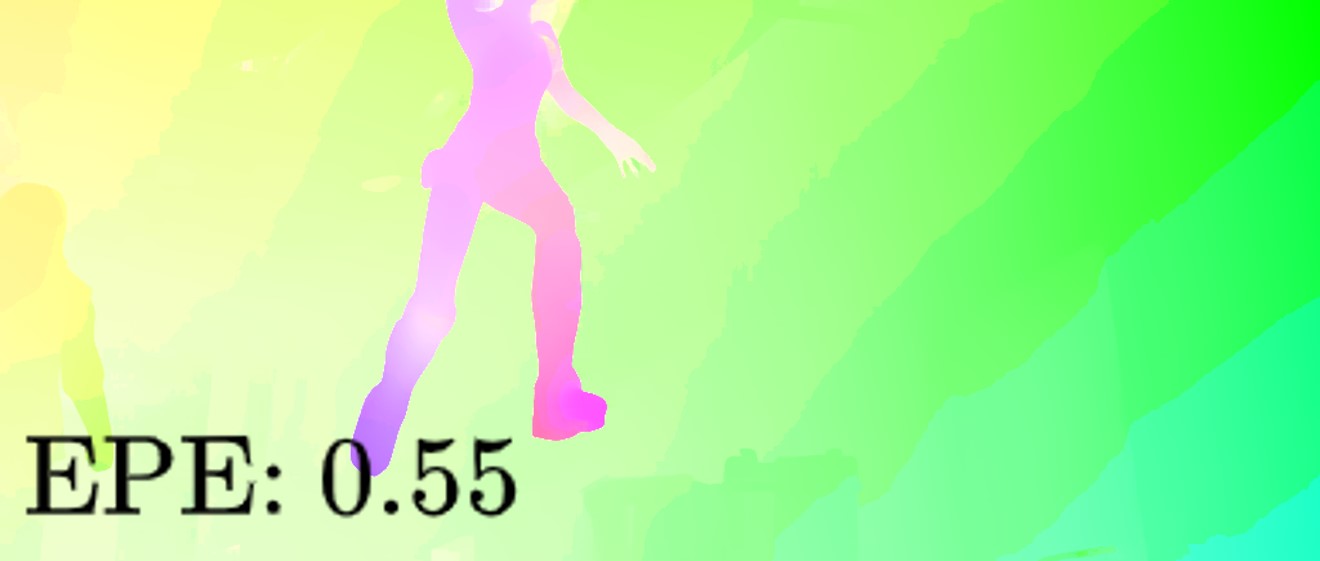}};
\spy on (-0.35,0.5) in node [left] at (2.65,0.0);\end{tikzpicture}
 & \hspace{-6mm} 
  \begin{tikzpicture}[spy using outlines={rectangle,red,magnification=3, size=1.6cm, connect spies}] \node {\includegraphics[trim=0 0 0 0, clip, width=3.7cm, trim={0 0 0 0},clip]{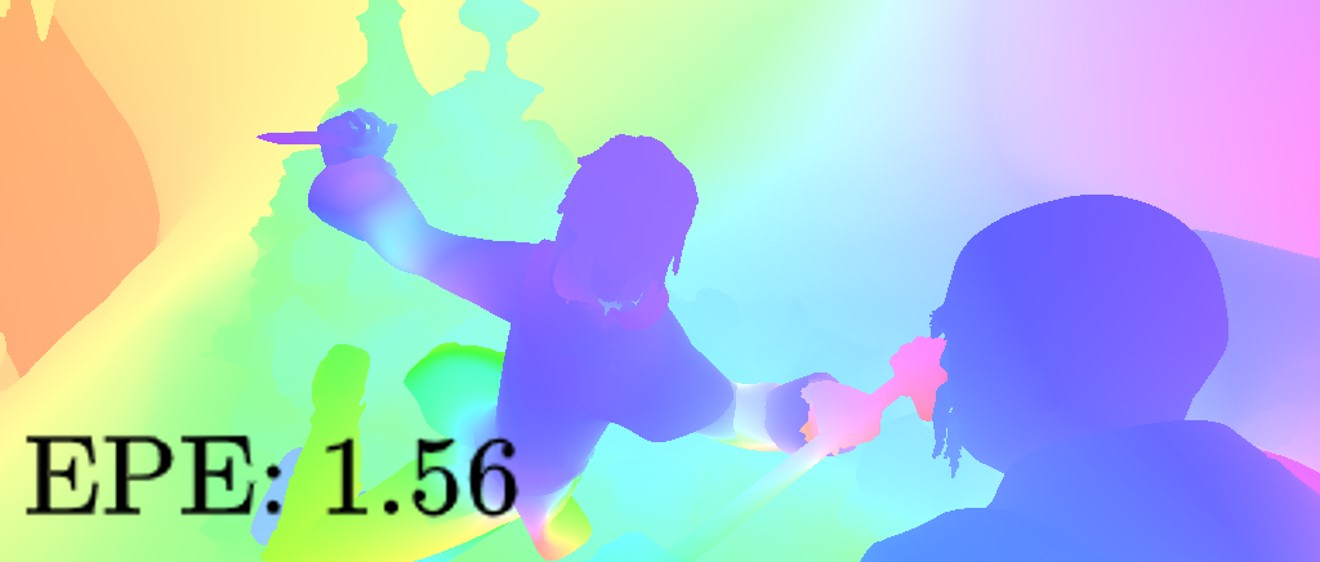}};
\spy on (-0.9,0.4) in node [left] at (2.65,0.0);\end{tikzpicture}
 & \hspace{-6mm} 
  \begin{tikzpicture}[spy using outlines={rectangle,red,magnification=3, size=1.6cm, connect spies}] \node {\includegraphics[trim=0 0 0 0, clip, width=3.7cm, trim={0 0 0 0},clip]{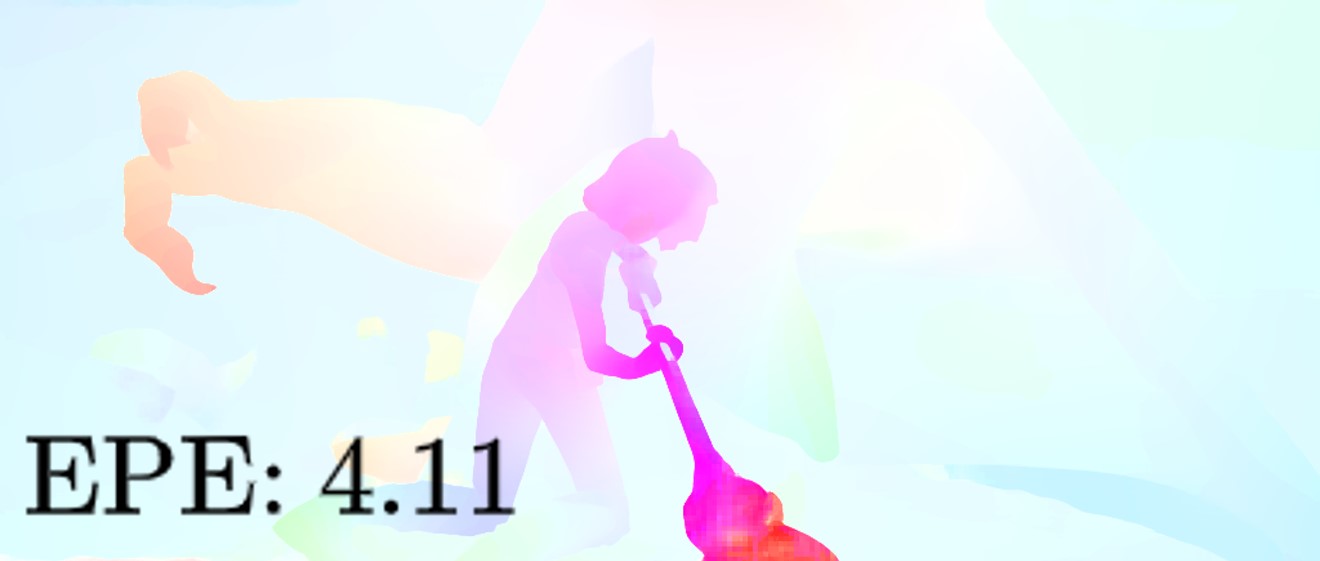}};
\spy on (0.3,-0.5) in node [left] at (2.65,0.0);\end{tikzpicture}
 \\
 
 \end{tabular}$
 \end{center}
 \caption{Qualitative results on Sintel (train) using GMFlow and NODE-GMFlow. The top row depicts the two frames overlay, the second row is a ground truth optical flow. We additionally zoom into regions highlighting the difference between these two models.}
 \label{fig:qualitative}
 \end{figure*}
 
 In this section, we present a comprehensive evaluation of \ours~on realistic or real-world data, including Sintel \cite{butler2012naturalistic} and KITTI \cite{geiger2013vision, menze2015object}.

\subsubsection*{Datasets}
We follow commonly used training and evaluation protocols~\cite{teed2020raft, jiang2021learning, zhang2021separable, huang2022flowformer, xu2023unifying}. 
First, we train our model on FlyingChairs (Chairs)~\cite{dosovitskiy2015flownet} and FlyingThings3D (Things)~\cite{mayer2016large} datasets. Using the model checkpoint after the first training stage, we evaluate the zero-shot generalization on Sintel (train) \cite{butler2012naturalistic} and KITTI (train) \cite{geiger2013vision, menze2015object} datasets. After that, we finetune the model on the mix of the datasets consisting of Sintel, KITTI and HD1K datasets. Following the training protocol of GMFlow, we additionally finetune the C + T checkpoint on Virtual KITTI dataset. 


\subsubsection*{Evaluation Metrics} 
We follow the widely used evaluation metrics for optical flow estimation. End-Point Error is an Euclidean distance between ground truth and flow prediction and Fl is the percentage of optical flow outlier. 


\subsubsection*{Networks and Training}
We use GMFlow \footnote{\url{https://github.com/autonomousvision/unimatch}} architecture as the baseline. The refinement module based on Convolutional GRU was replaced with the Neural ODE block. In our implementation, we used the codebase proposed in the original Neural ODE paper \cite{chen2018neural} \footnote{\url{https://github.com/rtqichen/torchdiffeq}}. The mixing network $\mathcal{M}$ consists of two convolutional layers with ReLU non-linearity mapping from 
$\mathbb{R}^{d_{inp} + d_{inp} + d_{hid}}\to \mathbb{R}^{d_{hid} + d_{hid}}$, with kernel size $5$ and padding $3$ and from $\mathbb{R}^{d_{hid} + d_{hid}}$ to $\mathbb{R}^{d_{hid}}$ with the same kernel size and padding. The tolerance parameter in the neural ODE block is set to $1e-3$.

\subsubsection*{Implementation details} Due to computational restrictions, we could not run the experiments with the hyperparameters from original GMFlow paper and had to scale the batch size and learning rates accordingly. We trained our model on FlyingChairs dataset for $50K$ iterations with a batch size of $32$ and learning rate of $4e-4$. The input images were resized to $(384, 512)$. Later, we finetuned the model on Things dataset for $800K$ iterations with a batch size equal to $8$ and learning rate of $1e-4$. Here, the input image size was set to be equal to $(384, 768)$. For later stages of training, we finetuned the model on Sintel dataset for $200K$ iterations with batch size of $4$ and learning rate of $1e-4$. For finetuning on Virtual KITTI, we initialized the model with the Things checkpoint and trained on Virtual KITTI dataset for $80K$ iterations with batch size of $8$ and learning rate of $5e-5$. Finally, finetuning on KITTI was done by training the model initialized with the Virtual KITTI phase checkpoint for $30K$ steps with batch size of $8$ and learning rate of $1e-4$. All models were trained using AdamW optimizer \cite{loshchilov2017decoupled} and with OneCycleLR scheduler \cite{smith2019super}.

\begin{table*}[t!]
\begin{center}
\caption{Optical flow estimation results on Sintel (train), and KITTI (train) datasets. We train the models on FlyingChairs (C) and FlyingThings (T). Similarly to the training protocol of GMFlow, we additionally train our model on Virtual KITTI dataset. Number in brackets (columns of KITTI in lower three rows of table) indicate results of models additionally trained on VK. We provide two options of our models: version (I) was trained for 400K iterations on FlyingThings, while (II) was trained for 800K iterations.} 
\vspace{3mm}
\label{tab:all_results}
{
\begin{tabular}{lcccccc}
\toprule
\multirow{2}{*}{Method} & Training & \multirow{2}{*}{$\#$Parameters} & \multicolumn{2}{c}{Sintel (train)} & \multicolumn{2}{c}{KITTI (train)}\\
\cmidrule{4-7}
& Dataset & & Clean & Final & EPE & Fl-all \\
\midrule[0.3pt] \midrule[0.3pt]
LiteFlowNet \cite{hui2018liteflownet} & \multirow{6}{*}{C+T} & 5.4M & 2.48 & 4.04 & 10.39 & 28.5 \\
FlowNet2 \cite{ilg2017flownet} & & 162.5M &  2.02 & 3.54 & 10.08 & 30.0 \\
PWC-Net \cite{sun2018pwc} & & 8.8M &  2.55 & 3.93 & 10.35 & 33.7 \\
RAFT \cite{teed2020raft} & & 5.3M &  1.43 & 2.71 & 5.04 & 17.4 \\
GMA \cite{jiang2021learning} & & 5.9M &  1.30 & 2.74 & 4.69 & 17.1\\
FlowFormer \cite{huang2022flowformer} & & 18.2M & \textbf{1.01} & 2.40 & 4.09 & 14.7 \\
\midrule
GMFlow \cite{zhao2022global} & \multirow{3}{*}{C+T / (+VK)} & 4.7M &  1.08 & 2.48 & 7.77 / (2.85) & 23.4 / (10.8) \\
NODE-GMFlow I (Ours) & & 9.2M &  1.04 & \textbf{2.39} & 7.21 / (2.44) & 22.2 / (8.8) \\
NODE-GMFlow II (Ours) & & 9.2M & \textbf{1.01} & 2.59 & 6.53 / (\textbf{2.38}) & 20.5 / (\textbf{8.4}) \\
\bottomrule
\end{tabular}
}
\end{center}
\end{table*}

\begin{table}[h!]
\centering
\caption{Optical flow estimation results on Sintel (test) and KITTI (test) datasets. For evaluation on test dataset, we used NODE-GMFlow II version. We observe a significant improvement over the baseline model GMFlow.
}
\label{tab:kittitest}
\vspace{2mm}
\begin{tabular}{lccc}
\toprule
\multirow{2}{*}{Method} & \multicolumn{2}{c}{Sintel (test)} & KITTI (test) \\
& Clean & Final & (Fl-all)\\
\hline \hline
LiteFlowNet2 \cite{hui2020lightweight} &3.48 & 4.69& 7.74 \\
PWC-Net+ \cite{sun2019models} & 3.45& 4.60& 7.72  \\
RAFT \cite{teed2020raft} &1.94 &3.18 & 5.10 \\
GMA \cite{jiang2021learning} & 1.40& 2.88& 5.15 \\
FlowFormer \cite{huang2022flowformer} & 1.16& 2.09&  4.68 \\
GMFlow+ \cite{xu2023unifying} & 1.03 & 2.37 & 4.49 \\
\hline
GMFlow \cite{zhao2022global} & 1.74 & 2.90 & 9.32 \\
NODE-GMFlow (Ours) & 1.26 & 2.55 & 5.41 \\
\bottomrule
\end{tabular}
\end{table}

\subsection{Evaluations on Sintel (train) and KITTI (train)}
Table~\ref{tab:all_results} shows the performance of previous state-of-the-art optical flow models and ours on Sintel (train) and KITTI (train). 
Our models trained on FlyingChairs and FlyingThings3D (C+T) shows significant improvements compared to GMFlow and comparable performance with state-of-the-art architectures on Sintel dataset. On KITTI dataset, although GMFlow and Ours show slightly worse performance than SOTA methods, our NODE-GMFlow models improve the performance over the baseline GMFlow model. Same as GMFlow, we additionally train Virtual KITTI 2 dataset, and our NODE-GMFlow shows the state-of-the-art performance on KITTI dataset.
While, in general, version (I) has better performance compared to GMFlow, training for more iterations seems to be more beneficial except for less accurate predictions on Sintel (final). More discussion on that can be found in Ablation section \ref{sec:ablation}.

\subsection{Evaluations on Sintel (test) and KITTI (test). }

\Cref{tab:kittitest} summarizes the performance of the existing methods and the accuracy of the proposed approach. Our model shows significant improvement compared to the baseline GMFlow model on both Sintel and KITTI datasets.



\subsection{ Qualitative Results}
\label{exp:qualitative}

 \begin{figure*}[h!]
 \begin{center}$
 \centering
 \begin{tabular}{c c}
 
 \hspace{0mm} \includegraphics[width=6.5cm]{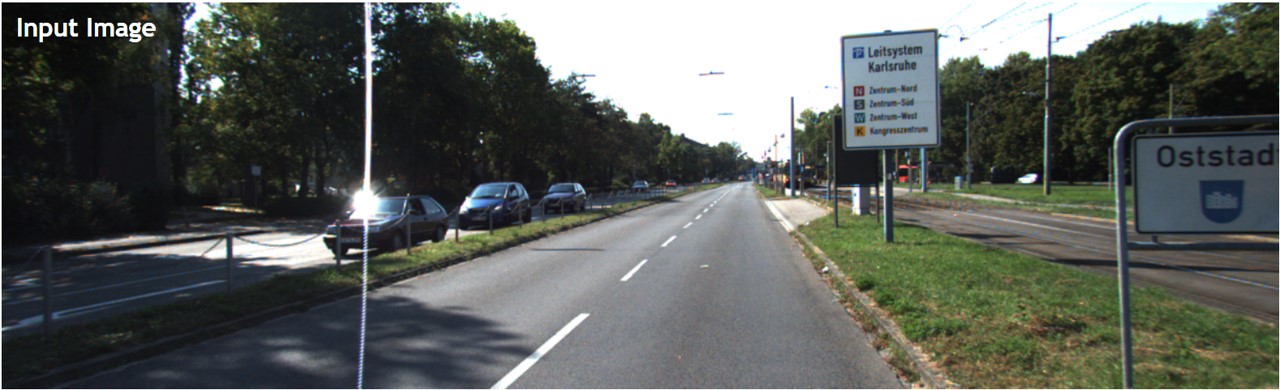} & 
 \hspace{0mm} \includegraphics[width=6.5cm]{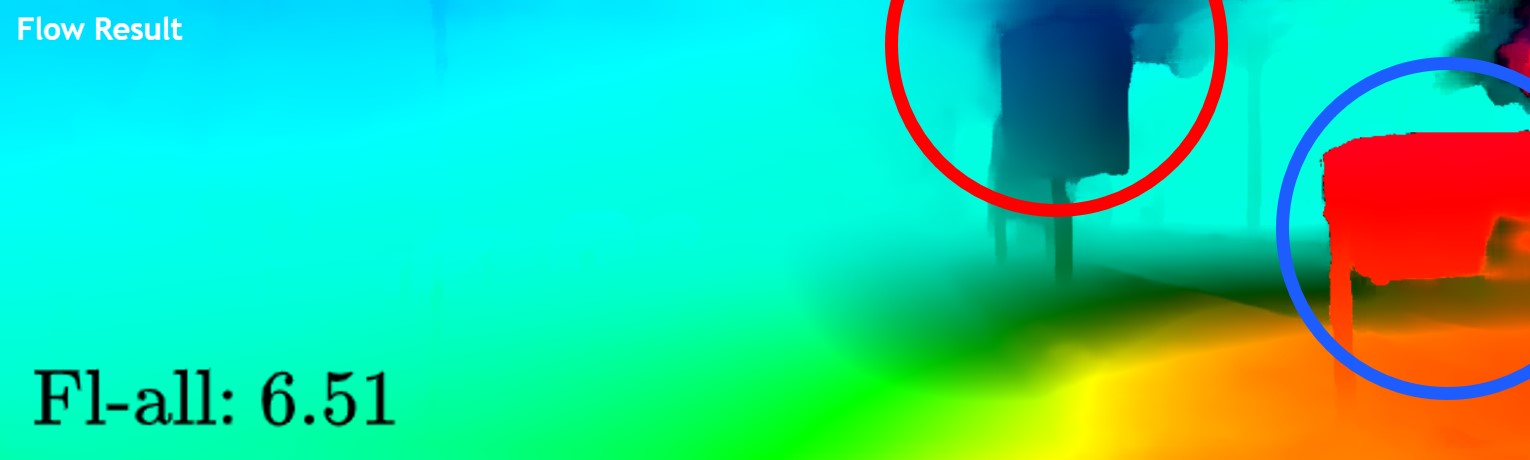} \\
 \text{(a) Image} &  \text{(b) GMFlow} \\
 \hspace{0mm} \includegraphics[width=6.5cm]{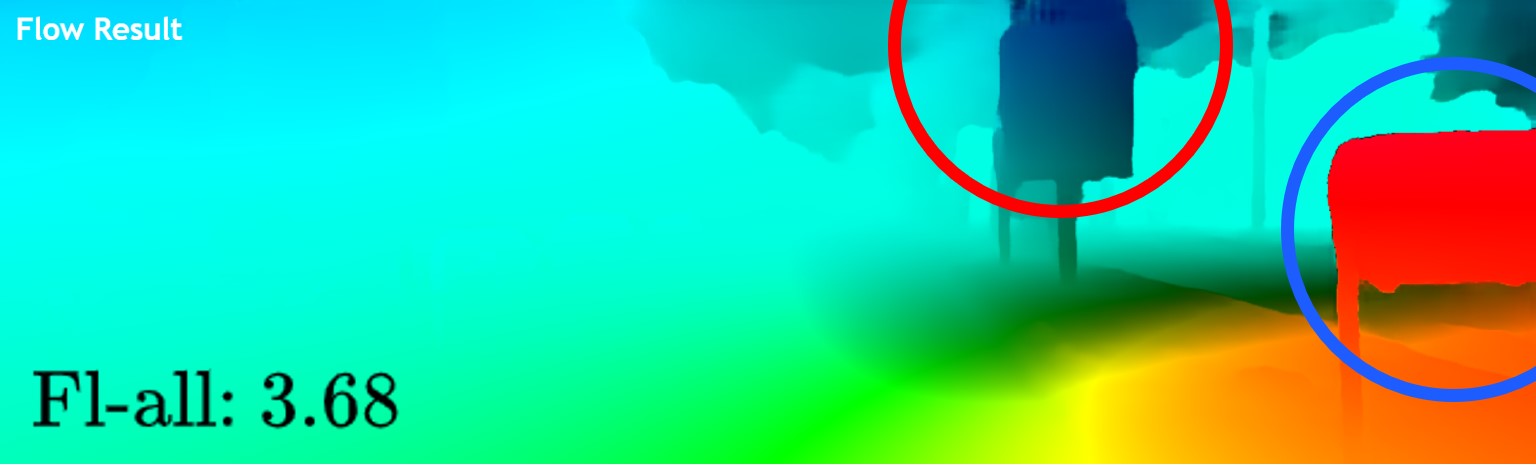} &
 \hspace{0mm} \includegraphics[width=6.5cm]{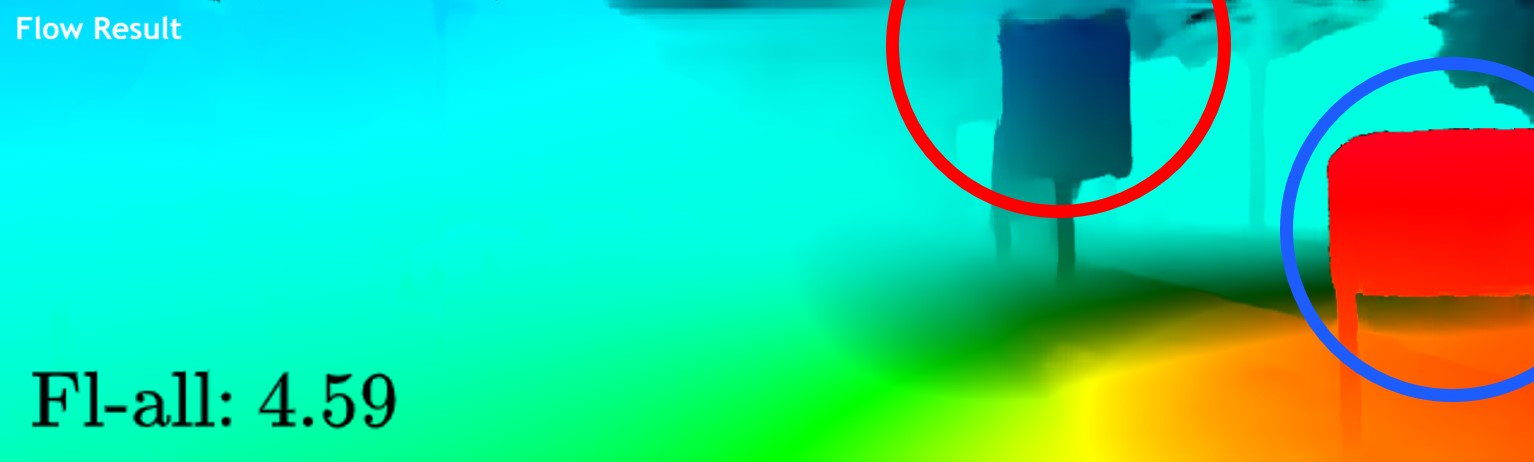} \\
  \text{(c) GMFlow+} &  \text{(d) NODE-GMFlow} \\

 \end{tabular}$
 \end{center}
 \caption{Qualitative results on KITTI (test) using GMFlow, GMFlow+, and NODE-GMFlow. Our model shows a significant improvement over the baseline GMFlow model with only one refinement step, while GMFlow+ uses 6 refinement steps on every iteration.}
 \label{fig:qualitative_kitti}
 \end{figure*}
Figure \ref{fig:qualitative} shows qualitative results on the Sintel (train, clean) using the original GMFlow and NODE-GMFlow which are trained on FlyingChairs and FlyingThings3D. Our NODE-GMFlow shows more accurate and robust flow estimation compared to GMFlow. 
Figure \ref{fig:qualitative_kitti} shows qualitative results on the KITTI (test) using GMFlow methods and NODE-GMFlow. Although our NODE-GMFlow fl-all score is slightly worse than GMFlow+, our NODE-GMFlow shows high quality prediction (as highlighted by red circle).



\subsection{Ablation Study}\label{sec:ablation}
In this part, we discuss the experimental results after varying parameters in the proposed model. We check the importance of the solver choice, number of training iterations and the structure of the mixing network.  


\begin{figure*}[h!]
\centering
\includegraphics[width=\textwidth]{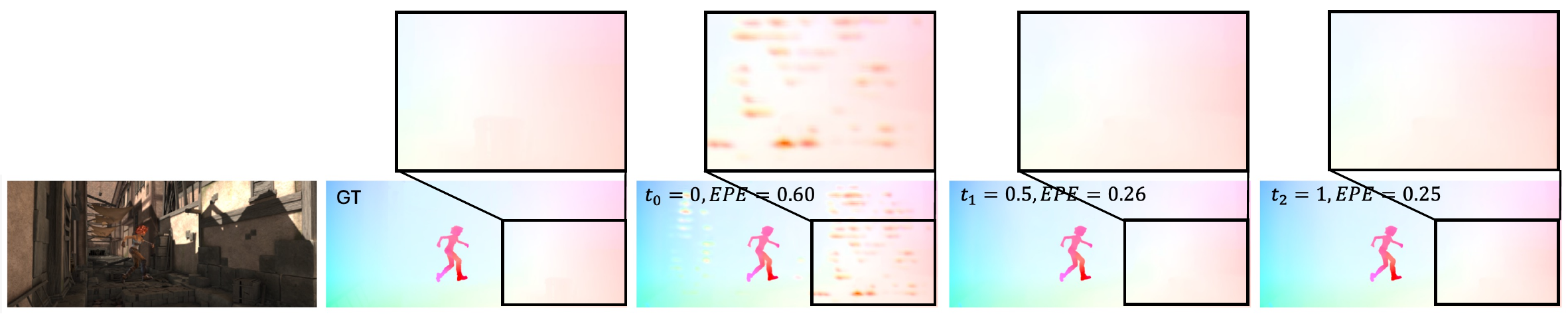}
\vspace{3mm}
\caption{
Instead of approximating the optical flow using iterative GRUs, we propose to use a Neural ODE that implicitly solves a differential equation. Neural ODEs model the flow in a continuous manner: the implicit integration is done from $0$ to $1$. In this figure, we visualize the convergence of the flow obtained from the intermediate steps during integration inside the Neural ODE block. As can be seen, the quality of the flow initially improves significantly.
}
\label{fig:flow_demonstration}
\end{figure*}
\subsubsection*{Varying time-scale $t$} The model was trained using boundary values of $t=0$ and $t=1$. In \Cref{fig:flow_demonstration} we demonstrate the effect of decoding the latent flow with time scales $t<1$. As can be seen, the quality improves significantly between $t=0$ and $t=0.5$, yet only marginally afterwards.

\subsubsection*{Mixing network} Before refining the predicted flow using the Neural ODE module, we project the input into the lower-dimensional space. Usually, this is done within the Convolutional GRU module mapping from $d_{inp} + d_{inp} + d_{hid}$ to $d_{hid}$. Instead, we first project the input to $\mathbb{R}^{d_{inp}}$ and then run Neural ODE module in the latent space. For the mixing network $\mathcal{M}$, we used two different networks: the first consisting of a single convolution and the second consisting of two  convolutional layers with ReLU in between. The experimental results are summarized in \Cref{tab:mixing}. We observe a better performance with the mixing network consisting of two convolutions. For our experiments, we choose this option.



\begin{table}[h!]
    \centering
    \caption{Number of iterations. Generally, an increased number of iterations leads to higher accuracy.}
    \label{tab:iters}
    \tabcolsep=0.11cm
    \begin{tabular}{lllllll}
        \hline 
        \multicolumn{1}{c}{\multirow{2}{*}{Model}} & \multicolumn{1}{c}{\multirow{2}{*}{Iters}} & Things & \multicolumn{2}{l}{Sintel (train)} & \multicolumn{2}{l}{KITTI (train)} \\ \cline{3-7} 
        \multicolumn{1}{c}{} &  & val        & clean            & final & EPE & Fl-all \\ \hline \hline
        NODE- \\ GMFlow (I) & 400K & 2.63 & 1.04 & \textbf{2.39} & 7.21 & 22.16 \\
        NODE- \\ GMFlow (II) & 800K & \textbf{2.47} & \textbf{1.01} & 2.59 & \textbf{6.52} & \textbf{20.46} \\ \hline
    \end{tabular}
\end{table}
\subsubsection*{Number of iterations}
We trained two models with different number of iterations, the experimental results are summarized in \Cref{tab:iters}. In general, the experiments suggest that training the model for more iterations benefits the prediction accuracy and generalization on unseen datasets. The original GMFlow model is trained for 800K iterations, however we can see the improvement over the baseline model even after training on 400K iterations; the comparison with the baseline can be found in the \Cref{tab:all_results}. For qualitative examples and final evaluations on KITTI dataset, we use the model trained for 800K iterations.

\begin{table}[!h]
    \centering
    \caption{Solver choice. All models were trained for 200K iterations on the Things dataset. }
    \label{tab:solver}   
    \begin{tabular}{llllll}
    \toprule
    \multicolumn{1}{c}{\multirow{2}{*}{Solver}} & Things & \multicolumn{2}{l}{Sintel (train)} & \multicolumn{2}{l}{KITTI (train)} \\ 
    \cmidrule{2-6}
    \multicolumn{1}{c}{}                        & val        & clean            & final           & EPE             & Fl-all              \\ 
    \midrule[0.3pt] \midrule[0.3pt]
    Fehlberg                                    & 2.88         & \textbf{1.14}             & \textbf{2.76}            & \textbf{7.20}            & 23.17           \\
    Midpoint                                    & \textbf{2.86}         & 1.15             & \textbf{2.76}            & 7.66            & \textbf{22.33}           \\ 
    \bottomrule
    \end{tabular}     
\end{table}

\subsubsection*{Solver}
We also experiment with different solvers; our findings are summarized in \Cref{tab:solver}. We can observe an almost identical accuracy on Things and Sintel, while on KITTI dataset the model trained with Fehlberg solver behaves slightly better. In later experiments, we resort to Midpoint solver due to better speed-accuracy tradeoff.

\begin{table}[!h]
    \centering
    \caption{Mixing network. Models were trained for 800K iterations. Choosing a deeper mixing network leads to a significantly increased accuracy.}
    \label{tab:mixing}
    \begin{tabular}{llllll}
    \toprule
    \multicolumn{1}{c}{\multirow{2}{*}{Mixing }} & Things & \multicolumn{2}{l}{Sintel (train)} & \multicolumn{2}{l}{KITTI (train)} \\ 
    \cmidrule{2-6}
    \multicolumn{1}{c}{network} & val & clean & final & EPE & Fl-all \\ 
    \midrule[0.3pt] \midrule[0.3pt]
    \begin{tabular}[c]{@{}l@{}}One conv \\ layer\end{tabular} & 2.55 & 1.06 & \textbf{2.59} & 6.71 & 21.50 \\
    \begin{tabular}[c]{@{}l@{}}Two conv \\ layers\end{tabular} & \textbf{2.47} & \textbf{1.01} & \textbf{2.59} & \textbf{6.52} & \textbf{20.46} \\ 
    \bottomrule
    \end{tabular}
\end{table}

\begin{figure*}[h!]
    \centering
    \begin{varwidth}{0.9\linewidth}
    \begin{subfigure}[t]{0.32\textwidth}
        \centering
        \includegraphics[width=\textwidth]{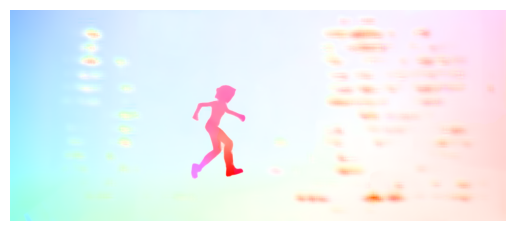}
        \caption{$t=-10$}
    \end{subfigure} \hfill
    \begin{subfigure}[t]{0.32\textwidth}
        \centering
        \includegraphics[width=\textwidth]{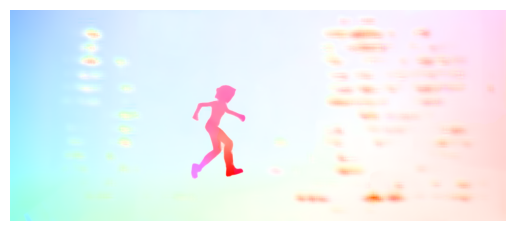}
        \caption{$t=-2$}
    \end{subfigure} \hfill
    \begin{subfigure}[t]{0.32\textwidth}
        \centering
        \includegraphics[width=\textwidth]{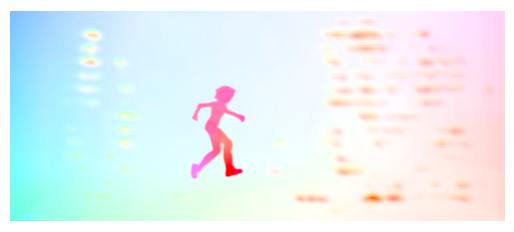}
        \caption{$t=0$}
    \end{subfigure} \hfill \\
    \begin{subfigure}[t]{0.32\textwidth}
        \centering
        \includegraphics[width=1.0\textwidth]{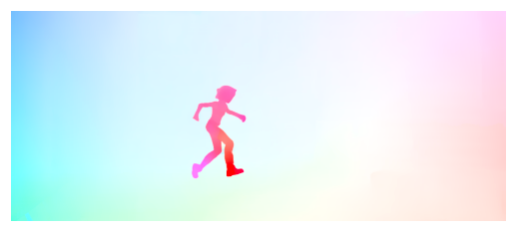}
        \caption{$t=1$}
    \end{subfigure}
    \begin{subfigure}[t]{0.32\textwidth}
        \centering
        \includegraphics[width=1.0\textwidth]{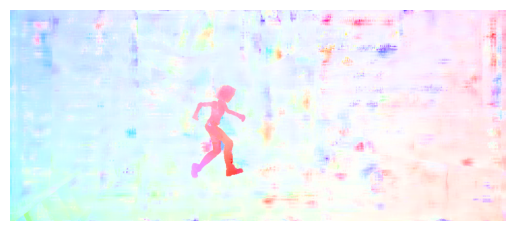}
        \caption{$t=3$}
    \end{subfigure}
    \begin{subfigure}[t]{0.32\textwidth}
        \centering
        \includegraphics[width=1.0\textwidth]{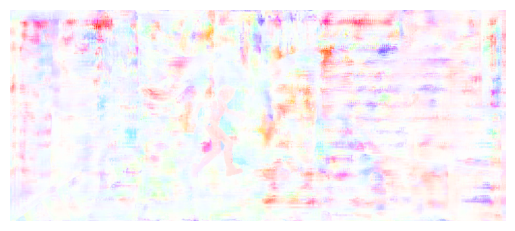}
        \caption{$t=5$}
    \end{subfigure}
    \end{varwidth}
    \caption{Extrapolating beyond the interval used for time scale $t \in [0,1]$ used during training. Using negative values for $t$ has little effect, yet the quality of the decoded flow quickly deteriorates for $t>1$.}
    \label{fig:extrapolation}
\end{figure*}

\subsubsection*{Extrapolating $t$}
To ablate the impact of the used time scale parameter in the neural ODE, we generated flows using time scales outside the training range. Those results are shown in \Cref{fig:extrapolation}. As can be seen extrapolating into the negative range has little effect compared with $t=0$. Extrapolating beyond $t=1$, however, causes the flow to become unreasonable.

\section{Limitations}\label{sec:limitations}
The limitations of the proposed method are the following. First, we observe a worse accuracy on Sintel final dataset thus our method is not improving the baseline model on all the benchmarks. Second, Neural ODEs are known to be prone to instabilities during the training~\cite{zhuang2020adaptive}. While we did not observe it during our experiments, there is the possibility of such problems if trained on other data. 
Finally, the proposed approach does not take into account novel advances in Neural ODE field, for instance~\cite{zhuang2020adaptive}.  

We believe that our work does not lead to negative societal impacts. The proposed approach is a foundational research and most of the training data is synthetic with the only real-life dataset used during the training being the KITTI dataset \cite{geiger2013vision, menze2015object}. On the contrary, we anticipate that our work impacts fields like autonomous driving which relies on OF estimation at the beginning of the information processing pipeline. As such, we hope that even small improvements in the accuracy of OF estimates may lead to fewer accidents involving self-driving vehicles.

\section{Conclusion}



In this paper, we proposed a novel approach that leverages Neural Ordinary Differential Equations to learn the optical flow field, based on given correlation and context features. Prior works use recurrent neural modules for refinement which are motivated from solvers in classical optimization but which require a fixed and pre-determined number of steps. The proposed method is more general than existing methods since it can generate the exact same updates if restrictive choices are made for its architecture, solver and hyperparameters. In other words, the proposed model can learn more general dynamics than works using recurrent neural modules. Extensive experiment results demonstrate the efficacy of our solution.

{
    \small
    \bibliographystyle{ieeenat_fullname}
    \bibliography{main}
}

\end{document}

%% file: method.tex

Many recent optical flow models adopt roughly the same pipeline for motion prediction. In this section, we provide a brief overview of a general setup and later describe our approach in more detail.

\subsection{Preliminaries}

Optical flow methods predict the motion displacement on a pixel level. Specifically, an optical flow $\mathbf{f}$ maps every position $\bbx = (x,y,t)$ at time $t$ to a position $\bbx^\prime = \bbx + \mathbf{f}(\bbx)$ at time $t+\tau$ by assuming brightness constancy
\begin{equation}
    \mathbf{f}(\bbx) = \mathbf{f}(\bbx + \Delta \bbx).
    \label{eq:brightness_constancy}
\end{equation}

Recently proposed methods usually tackle this problem in the following manner. Given a pair of images $\mathbf{I_1},\mathbf{I_2} \in \mathbb{R}^{H \times W \times 3}$ that are two consecutive video frames, the model first extracts low-dimensional representations $\bbg_1, \bbg_2 \in \mathbb{R}^{ H/n \times W/n \times D}$, where $D$ denotes the hidden feature dimension and $n$ usually equals $8$ (e.g., in RAFT ~\cite{teed2020raft}, FlowFormer ~\cite{huang2022flowformer}, GMFlow ~\cite{xu2022gmflow}). Sometimes an additional context network extracting the features only from the first image $\mathbf{I_1}$ is introduced \cite{teed2020raft, huang2022flowformer, xu2022gmflow}. Effectively, the context network is identical to the feature encoder network. 

Since it is expected for correspondences in the consecutive frames to have a high similarity, the next step is usually a construction of a correlation volume. In practice, it is realized as a four-dimensional tensor $\mathcal{C} \in \mathbb{R}^{H\times W \times H \times W}$ with entries corresponding to the dot product between the corresponding feature vectors. By pooling across two last dimensions, we obtain a correlation pyramid $\{\mathcal{C}_1, \ldots, \mathcal{C}_n \}$ with $\mathcal{C}_i \in \mathbb{R}^{H \times W \times H / {2^k} \times W / {2^k}}$, where $k$ usually runs from $0$ to $3$ ~\cite{teed2020raft}. 

One of the main stages in existing optical flow models is the refinement step. Integrating this additional stage into models helps to achieve a higher accuracy. Usually, the refinement is done by iteratively applying GRU-based updates.
In this setting, we enforce a \emph{discrete} nature of the refinement. Instead, we propose to generalize it to the \emph{continuous} case by replacing GRU with a Neural ODE block. Moreover, GRU-based update may be rewritten in a Neural ODE setting making it a more general setting.

The first flow approximation is initialized as zero $\bbf_0 = \mathbf{0}$. Then, the current flow estimation, a correlation and a context embedding $\mathbf{q}$ are fed into the iterative update module. Typically, the update module is realized as a separable convolutional GRU network and its output $\Delta \bbf$ is fed again into the update module several times in order to mimic the optimization process: $\bbf_{k +1}= \bbf_k + \Delta \bbf$. 
In this paper, we suggest to go beyond beyond ConvGRUs as the primary building block for optical flow refinement. Instead, we introduce a Neural ODE-based update module.

\subsection{Neural Ordinary Differential Equations}


Neural ODEs reformulate recurrent updates in the continuous manner. Suppose we have an update equation given by:
\begin{equation}
    \mathbf{h}_{t+1} = \mathbf{h}_t + g(\mathbf{h}_t,\theta_t),
\end{equation} where $t \in \{0,\ldots,T\}$ is a time step, $\mathbf{h}_t \in \mathbb{R}^n$ is a hidden state, $\theta$ is for model parameters and $g(\cdot)$ is a dynamics generating function. The continuous version of this dynamics can be written as:
\begin{equation}\label{eq:ode}
    \frac{d\mathbf{h}(t)}{dt} = g(\mathbf{h}(t),t,\theta_t).
\end{equation}

As proposed in \cite{chen2018neural}, we parameterize the displacement field by a neural network $f_\theta$ with learnable parameters $\theta$. 

In practice, the right-hand side of the equation \ref{eq:ode} is parametrized by a neural network. In essence, instead of explicitly parametrizing the underlying function itself, we parametrize the \emph{change} of the function. During Neural ODE training, the output is calculated using black-box differential equation solvers. By choosing the adjoint sensitivity method, we can use memory-efficient $O(1)$ backpropagation for updating the model parameters $\theta$ \cite{pontryagin2018mathematical}. 

\subsection{Parametrizing the right-hand side of the Neural ODE}
In optical flow models the flow estimation $\mathbf{f}$ is usually iteratively updated as $\mathbf{f}_{k+1} = \mathbf{f}_{k} + \Delta \mathbf{f}$. In this paper, we propose to model $\Delta \mathbf{f}$ as an output of the neural ODE block. 






\subsubsection*{Transformer right-hand side}
Since we do not limit ourselves to the specific type of the architecture for the right-hand side of the ODE, an alternative option is to use a transformer-based right-hand side. This allows to model more global dependencies due to increased receptive field. In our experiments, we considered a simple transformer block. Since the dynamics modeled by ODE has to have the same output dimension as the input, we apply an additional module that projects the concatenated flow and correlation. In our experiments, we use a simple convolutional network consisting either of only one convolutional layer or of two convolutions with ReLU nonlinearity in between. More details on the architecture can be found in the next sections. 

Additionally, since we model the \emph{gradient} of the flow update, the actual solution is of a higher order, thus having a higher representation power. This property allows us to keep the right-hand side of the ODE very simplistic: we use only one Transformer block with one head and hidden MLP dimension the same as the input dimension. Besides that, we have a mixing network projecting the contatenated input to a hidden dimension represented by only one convolutional layer. However, this basic setup allows to model the optical flow with a higher accuracy compared to convolutional GRUs. 


\subsection{Neural ODE solvers}
Neural ODEs are being solved using black-box differential solvers. We have a flexibility when setting a solver: one can choose an adaptive step solver or a solver allowing for setting the step size and a maximal number of iterations allowed for a solver. Apart from that, one can choose between higher-order solvers (as some of the Runge-Kutta methods \cite{runge1895numerische, kutta1901beitrag}) or first or second order solvers (e.g., Euler and midpoint methods, respectively). In our experiments, we chose a simple midpoint method that is a refinement of the Euler method. We observed a reasonable tradeoff between the convergence speed and accuracy but one can always adjust the solver choice and tolerance parameters. In some experiments, we have used Fehlberg method, which is an adaptive fourth-order method from Runge-Kutta family. We did not notice a significant difference in accuracy while varying a solver, thus, in later experiments we just chose a faster option.

\subsection{Training Loss}
We adopt the same training losses as in the baseline model: 
\begin{equation}
    \mathcal{L} = \sum_{i=1}^N \gamma^{N-i}\|\mathbf{f}_{gt} - \mathbf{f}_i\|_1,
\end{equation}
where $\mathbf{f}_{gt}$ is a ground truth flow, $\mathbf{f}_i$ is a predicted flow on the $i$-th timestamp, $N$ is a number of predictions and $\gamma=0.9$ is a weighting coefficient. In our experiments, we use only 1 refinement step.

\subsection{Architecture details} For the Neural ODE right-hand side block we use a Transformer block with input and output dim $d_{inp} = d_{out} = 128$, the hidden dimension in the MLP layer equals $d_{hid}=128$, number of heads equals to one. The mixing network is a mapping $$\mathcal{M}\big([\mathbf{q}, \bbf, \mathcal{C}]\big) \colon \mathbb{R}^{d_{inp}} \times \mathbb{R}^{d_{inp}} \times \mathbb{R}^{d_{hid}} \to \mathbb{R}^{d_{hid}},$$ where $\mathbf{q}$ is a context embedding, $\bbf$ is a flow estimation and  $\mathcal{C}$ is a correlation tensor. We use two versions: one convolutional layer or a small convolutional network consisting of two layers.